\documentclass[10pt,twocolumn,letterpaper]{article}

\usepackage[pagenumbers]{cvpr} % To force page numbers, e.g. for an arXiv version

\usepackage{graphicx}
\usepackage{amsmath}
\usepackage{amssymb}
\usepackage{pifont}
\usepackage{booktabs}

\makeatletter
\@namedef{ver@everyshi.sty}{}
\makeatother
\usepackage{tikz}
\usepackage{pgfplots}
\usepackage{multirow}
\usetikzlibrary{arrows}
\usepackage{enumitem}
\setlist{nosep}
\usepackage[pagebackref,breaklinks,colorlinks]{hyperref}

\usepackage[capitalize]{cleveref}
\crefname{section}{Sec.}{Secs.}
\Crefname{section}{Section}{Sections}
\Crefname{table}{Table}{Tables}
\crefname{table}{Tab.}{Tabs.}
\newcommand{\inv}{\mathit{inv}}
\newcommand{\cyc}{\mathit{cyc}}
\newcommand{\Id}{\mathit{Id}}
\newcommand{\GAN}{\mathit{GAN}}
\newcommand{\cmark}{\ding{51}}%
\newcommand{\xmark}{\ding{55}}%

\newcommand*{\affmark}[1][*]{\textsuperscript{#1}}
\def\cvprPaperID{10098} % *** Enter the CVPR Paper ID here
\def\confName{CVPR}
\def\confYear{2023}

\begin{document}

\title{ParGANDA: Making Synthetic Pedestrians A Reality For Object Detection}

% \and

% \author{
% Shuzhi Yu\affmark[1]\thanks{This work was done when Shuzhi Yu was an intern at Google} \;\;\;\;
% Guanhang Wu\affmark[2] \;\;\;\;
% Chunhui Gu\affmark[2] \;\;\;\;
% Mohammed E. Fathy\affmark[2] \\
% \affmark[1]Duke University \;\;\;\; \affmark[2] Google LLC \\
% {\tt\small shuzhiyu@cs.duke.edu} \;\;\;\; {\tt\small \{guanhangwu,chunhui,msalem\}@google.com}
\author{
Daria Reshetova\affmark[1]\thanks{This work was done when Daria Reshetova was an intern at Google} \;\;\;\;
Guanhang Wu\affmark[2] \;\;\;\;
Marcel Puyat\affmark[2] \;\;\;\;
Chunhui Gu\affmark[2] \;\;\;\;
Huizhong Chen\affmark[2] \\
\affmark[1]Stanford University \;\;\;\; \affmark[2] Google LLC \\
{\tt\small resh@stanford.edu} \;\;\;\; {\tt\small \{guanhangwu,marcelpuyat,chunhui,huizhongc\}@google.com}
}
% Daria Reshetova, guanhangwu@google.com, marcelpuyat@google.com, Chunhui Gu, Huizhong Chen
\maketitle

%%%%%%%%% ABSTRACT
\begin{abstract}
     Object detection is the key technique to a number of Computer Vision applications, but it often requires large amounts of annotated data to achieve decent results. Moreover, for pedestrian detection specifically, the collected data might contain some personally identifiable information (PII), which is highly restricted in many countries. This label intensive and privacy concerning task has recently led to an increasing interest in training the detection models using synthetically generated pedestrian datasets collected with a photo-realistic video game engine. The engine is able to generate unlimited amounts of data with precise and consistent annotations, which gives potential for significant gains in the real-world applications. However, the use of synthetic data for training introduces a synthetic-to-real domain shift aggravating the final performance. To close the gap between the real and synthetic data, we propose to use a Generative Adversarial Network (GAN), which performs
     %a modification of Parametric GAN for Domain Adaptation (ParGANDA), which performs 
     parameterized unpaired image-to-image translation to generate more realistic images.
     The key benefit of using the GAN is its intrinsic preference of low-level changes to geometric ones, which means annotations of a given synthetic image remain accurate even after domain translation is performed thus eliminating the need for labeling real data.
     We extensively experimented with the proposed method using MOTSynth dataset to train and MOT17 and MOT20 detection datasets to test, with experimental results demonstrating the effectiveness of this method. 
     Our approach not only produces visually plausible samples but also does not require any labels of the real domain thus making it applicable to the variety of downstream tasks.
\end{abstract}

     %that will be then used for training hence improving the performance in real applications. 
     %Finally, we demonstrate that the adaptation process generalizes to object classes unseen during training. In particular we show a significant improvement compared to training on raw synthetic data both with and without fine-tuning on real data.
%%%%%%%%% BODY TEXT
\section{Introduction}
\label{sec:intro}

% We need layers to draw the block diagram
\pgfdeclarelayer{background}
\pgfdeclarelayer{foreground}
\pgfsetlayers{background,main,foreground}

\begin{figure}[ht]
    \centering
    \begin{subfigure}[b]{\columnwidth}
    \includegraphics[width=\textwidth]{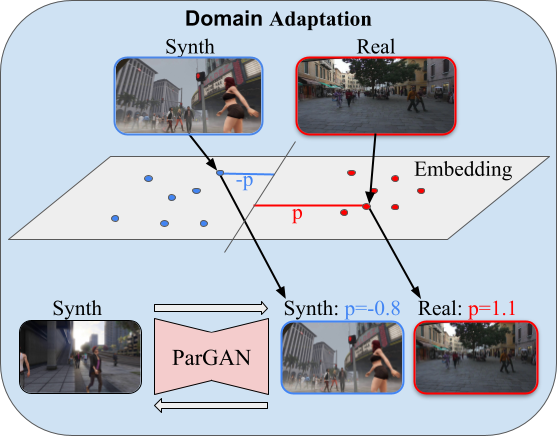}
    \caption{Phase 1: domain adaptation. The "realness" parameters are calculated for all the images in both the real and synthetic datasets and then used to train the ParGAN model.}
    \end{subfigure}
    \begin{subfigure}[b]{\columnwidth}
    \includegraphics[width=\textwidth]{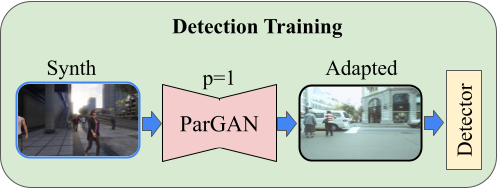}
    \caption{Phase 2: detection training. The learned ParGAN mapping is used as a preprocessor for the detector algorithm. The detection bounding boxes do not change between the synthetic and the output domains.}
    \end{subfigure}
    \caption{Two-phase training of the detection model with parametric domain adaptation}
    \label{fig:training_scheme}
\end{figure}

% \begin{figure*}[ht]
%     \centering
%     \includegraphics[width=\textwidth]{inference.png}
%     \caption{Examples of the ParGAN output -- the bottom images are the "more real" ones generated by the ParGAN from the top ones}
%     \label{fig:inference}
% \end{figure*}

\begin{figure*}[ht!]
\centering
\begin{tabular}{cc}
 \centering
      \raisebox{2em}{\rotatebox[origin=c]{90}{MOTSyn}} \hspace{-1em}&\hspace{-1em}\includegraphics[width=0.96\textwidth]{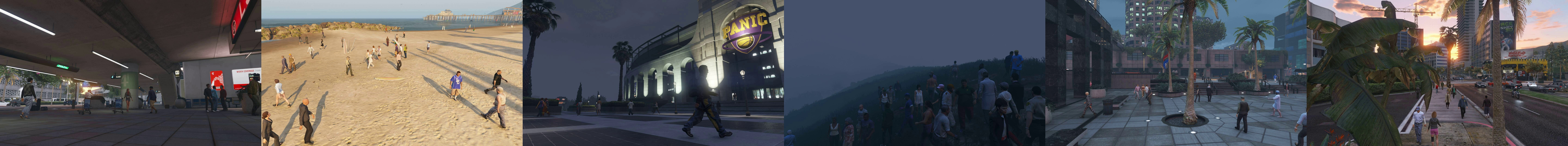}
      \vspace{-0.5em} \\
      \raisebox{2em}{\rotatebox[origin=c]{90}{ParGAN}}
      \hspace{-1em}&\hspace{-1.1em}\includegraphics[width=0.96\textwidth]{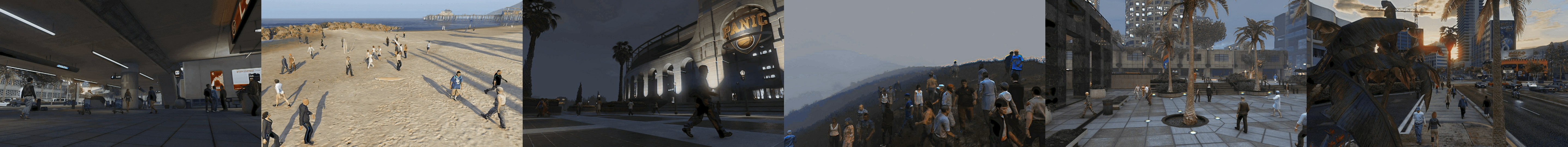}
 \end{tabular}
    \caption{Examples of the ParGAN output -- the bottom images are the "more real" ones generated by the ParGAN from the top ones}
    \label{fig:inference}
\end{figure*}

Well–annotated datasets for pedestrian detection such as  \cite{maggiori2017dataset,dollar2011pedestrian,zhang2017citypersons,milan2016mot16,maggiori2017can} are crucial
for the task. However, creating such datasets is not only expensive, but also comes with a number of legal challenges, which leads to the datasets being relatively small and/or having inconsistent labelling. Therefore, adapting synthetic data for model training is critical for pedestrian detection, perhaps even more so than in other computer vision applications. 

Advanced video-game engines seem to be a viable solution for this problem since they produce synthetic images of with precise consistent labels.
However, these synthetic images still come from a different underlying distribution than the real ones and can be easily distinguished from the real ones. Therefore, the models trained on such synthetic data experience a performance drop when evaluated on real images due to the difference in train and test distributions, which is commonly referred to as a domain gap.
% The existence of this domain gap is indicated by the difference in the performance of the model trained solely on the synthetic data versus the fine-tuned one as indicated in 

% A number of domain adaptation (DA) techniques can be used to address this problem, and in this work we focus on the framework of GANs primarily due to their ability to prioritize low-lever transformations \cite{dhariwal2021diffusion} over large geometric ones. For the object detection task, the lack of major geometric changes leads to the bounding boxes of the objects being the same for both the input and generated image, which makes the domain adaptation task not only significantly simpler but also allows for adaptation to unlabelled domains. While the preference of GAN-based models to perform rather small geometric changes might seem concerning in other tasks, our experiments show that the changes in the low-level features and minor geometric changes are enough for a dramatic increase in the performance of the detection model.

Several domain adaptation (DA) techniques offer promising solutions to reducing the domain gap. In this work, we concentrate predominantly on the framework of Generative Adversarial Networks (GANs), owing to their inherent capacity to emphasize low-level transformations over substantial geometric ones \cite{dhariwal2021diffusion}. When applied to the task of object detection, the absence of considerable geometric alterations leads to the invariance in the bounding boxes of objects across both the input and synthesized images. This particular characteristic not only simplifies the domain adaptation task but also paves the way for adaptation towards unlabeled domains. Although the propensity of GAN-based models to effectuate relatively minor geometric alterations may raise concerns in certain tasks, our experimental results validate that modifications to low-level features, coupled with minor geometric adjustments, can significantly enhance the performance of the detection model. Moreover, to the best of our knowledge, it is the first method for domain adaptation for pedestrian detection that does not require any labeled data from the real domain or use any real data for training the detection model, which is crucial for PII data.

% In this work, we use a parametric GAN (ParGAN) \cite{pargan} to perform DA. The method finds an invertible mapping from the source space of synthetic images to the target space of real images but also allows for an addition of a parameter that represents the "realness" score of the image, Intuitively, for real images the score is high and for synthetic ones, it is much lower. At the inference stage, any realness parameter can be passed  to the model together with a synthetic image to generate an image with a certain degree of "realness", see Fig. \ref{fig:training_scheme}. While it does not directly serve the final goal of generating real images, it helps make the DA problem better posed as it treats "realness" as a continuous spectrum as opposed to a binary property.

Based on the varying quality of the synthetic and real images, we propose to introduce a parameter signifying how real the image looks. We will call this the 'realness' score of an image. Intuitively, this score exhibits a higher value for real images and a substantially lower one for the synthetic ones, thus making 'realness' a spectrum as opposed to a binary property, thereby fostering a more nuanced understanding of the task. To incorporate this idea we use a parametric GAN (ParGAN) \cite{pargan} to facilitate domain adaptation. This method discerns an invertible mapping from the source domain of synthetic images towards the target domain of images (both synthetic and real) parametrized with the 'realness' score. During the inference phase, the model accepts any 'realness' parameter, along with a synthetic image, to produce an image with a specified degree of 'realness', as depicted in Fig. \ref{fig:training_scheme}. Even though this does not explicitly contribute to the ultimate objective of generating real images, it aids in refining the domain adaptation problem. Specifically, it delivers a more precise distinction between synthetic and real images, thereby reducing the model confusion between synthetic images that closely resemble real ones and truly real images.

% The main reason for choosing a GAN as the base for our domain adaptation method interestingly emerges from the long-standing observation that GANs tend to prioritize low-level transformations  \cite{dhariwal2021diffusion} 
% % criticism of the method. and the long-standing observation that they do not handle dramatic geometric changes well \cite{dhariwal2021diffusion}.
% For the object detection task, the lack of major geometric changes means that the bounding boxes of the objects are the same for both the input and generated image, which makes the domain adaptation task not only significantly simpler but also allows for adaptation to unlabelled domains. While the preference of ParGAN-like models to perform rather small geometric changes might seem concerning, our experiments show that the changes in the low-level features and minor geometric changes are enough for a dramatic increase in the performance of the detection model.

We also emphasise that while this paper focuses primarily on the detection problem, the separation of the domain adaptation from the downstream task makes our method applicable to a variety of applications when labels are based on the geometric features of the image, such as semantic segmentation or object tracking. 
In short, the main contributions of our paper are
\begin{itemize}
    \item a novel versatile method for synthetic-to-real domain adaptation that preserves the image geometry
    \item state-of-the-art results for detection on MOT17 without the use of any real labels
    \item a new way of parameterizing the data shift between the two domains --  the "realness" parameter of images that indicates how real the image is and improves the domain adaptation
\end{itemize}
% TODO come up with a third point

%-------------------------------------------------------------------------
% Stopped here
\section{Related Work}
% Object detection and pedestrian detection, in particular, is an active area of research [add citations \cite{pedestrian_detection}] with most of the state-of-the-art results are due to the refinement of the detector and employing the additional  .... 
It was recently shown \cite{fabbri2021motsynth,ciampi2020virtual} that labeled synthetic data from highly photo-realistic video-game engines can significantly improve the accuracy of pedestrian detection models.
%While these model and training algorithms do improve the performance of the final model, they require significant changes in the model architecture and/or training process. A somewhat orthogonal way to improve the final model involves enriching the training dataset.
%, and with the recent emergence of synthetic datasets labeled for pedestrian detection \cite{fabbri2021motsynth,ciampi2020virtual} it became possible to improve existing detection models with synthetic data. 
In this section, we provide a brief review of work related to synthetic data in pedestrian detection and the synthetic to real domain adaptation techniques.

% Domain adaptation is an active area of research [add citations here], however, the , 

\subsection{Synthetic Datasets for Pedestrian Detection}
High-quality, diverse, well-labeled datasets are crucial for successful detection, and while there exists a number of real-world datasets, such as INRIA \cite{maggiori2017dataset}, Caltech Pedestrian \cite{dollar2011pedestrian}, CityPersons \cite{zhang2017citypersons}, MOT17Det\cite{milan2016mot16} and MOT20 \cite{maggiori2017can}, they are comparatively small given the prohibitive costs of human-driven labeling and the legal constraints of data being non-personally identifiable. The size constraint poses a singular problem for the training of large-scale pedestrian detection models as they need vast amounts of data for precise detection.
%so generation of synthetic datasets with high precision rendered labels has been actively exploited \cite{fabbri2021motsynth,ciampi2020virtual}. 

As a result, there have been a lot of recent studies looking for cheaper ways to create big, accurate, and well-labeled collections of data. The pioneering work of  \cite{fabbri2018learning} focused on the task of multi-people tracking in urban scenarios, and proposed to add images generated by a highly photo-realistic video game engine with precise labels to solve the problem of occluded pedestrians, since it is impossible to get exact positions of people who are hidden in real data. The authors of \cite{fabbri2018learning} created a vast dataset of human body parts for people tracking in urban scenarios and showed that it can increase the accuracy of multi-people tracking and pose estimation.

Consequent works of \cite{ciampi2020virtual,fabbri2021motsynth} further employ the video-game engine to generate consistently and precisely labelled synthetic data. The video game engine provides a cheap and consistent way to create labelled data for pedestrian tracking/detection -- the larger dataset (\cite{fabbri2021motsynth}) contains more than 1 million frames with more than 40 million pedestrians compared to at most 50k frames and 1 million pedestrians for real datasets \cite{milan2016mot16,andriluka2018posetrack,dendorfer2020mot20}. The work of \cite{ciampi2020virtual} presented the first synthetic dataset suitable for pedestrian detection task, and while training the model solely on synthetic data  was comparable to training one on the real data, the model still experienced the synthetic to real domain shift indicated by the performance drop on synthetic vs real test data. The authors proposed 2 different techniques to reduce it: training on the synthetic data and fine-tuning on the real samples and mixed batch training, when batches from the two datasets were mixed at the training phase. While these methods gave reasonable performance increase, we would like to point out that they do require labels of the real dataset, which are not always accessible.

% for pedestrian detection demonstrate the successful use of labelled images generated by a video-game engine for pedestrian detection.  and report an increase in mean average precision (mAP) for detection models trained purely in synthetic data compared to training on real labels. 

The work of \cite{fabbri2021motsynth} developed a different synthetic dataset, which was also generated by the highly photo-realistic video-game engine. They showed, for example, that training only on synthetic data improves the detector accuracy by 10\% on MOT17 \cite{milan2016mot16} (no real labels are used) compared to training purely on real data. We use the dataset they developed (MOTSynth), which is a diverse synthetic dataset for pedestrian detection, segmentation and tracking. It consists of 768 videos of different indoor and outdoor scenarios in a variety of weather conditions. We use it and to perform domain adaptation followed by training the detection model on the domain adapted images.

\subsection{Synthetic-to-real domain adaptation}
% Moreover, for MOT17 our method outperforms the detector trained with the real labels and thus makes labeling real images obsolete.
While our method largely relies on an extensive synthetic dataset, an equally important building block is synthetic-to-real domain adaptation technique. Domain adaptation is an active field of research where deep neural networks show their empirical success \cite{ganin2016domain,long2015learning,sun2016return,tzeng2014deep,isola2017image,zhu2017unpaired}. 

Previous works that focus on using synthetic data for pedestrian detection either combine the samples from both domains \cite{ciampi2020virtual}, %\cite{vazquez2013virtual},\cite{li2014domain},
or generate pseudo-annotations for the target domain and train a detector based on those annotations  \cite{liu2016unsupervised}. However, this can be problematic when the target data is as scarce as in MOT17 \cite{milan2016mot16} and MOT20 \cite{dendorfer2020mot20} or is legally protected. Other works on domain adaptation require paired data, which is not available for synthetic-to-real domain transfer. This forces us to focus on domain adaptation for unpaired data. The techniques we explore define an image-to-image mapping that brings the input closer to the desired domain.

\textbf{Style transfer}
One of the pioneers of deep-learning-based domain adaptation in computer vision is Gatys et al. \cite{gatys2016image}, who introduced the image style transfer. Style transfer applies the style of one image to another while preserving its context. This is done by disentangling the content statistics from the style statistics of the two images and optimizing the output to match them.
Johnson et al. \cite{johnson2016perceptual} built on \cite{gatys2016image} to make style transfer work in real-time by creating style transfer networks. While this approach might at first seem right for our problem, both of these approaches transfer the style of one single image. For our task, we need to replicate the style of an entire domain of real images and do not want to lose variability in the translated dataset.

\textbf{Image-to-Image translation}
The above caveat of style transfer brings us to the more general problem of image-to-image translation. This framework is perhaps the most general way to perform domain adaptation, which consists of mapping an image from one domain to another and optimizing the map to minimize a chosen distance to the domain. GAN-based architectures like Pix2Pix \cite{isola2017image} learn
the mappings (black-and-white images $\leftrightarrow$ colored images, edges $\leftrightarrow$ photo, day $\leftrightarrow$ night) optimize the distance between the paired images in the two domains and achieve very realistic results.
However, in the case of synthetic pedestrians $\leftrightarrow$ real images, obtaining paired datasets is impossible, which brings us to unpaired image-to-image translation. This is known to be well-performed by cycle-consistent architectures, such as CycleGAN \cite{zhu2017unpaired}, DiscoGAN \cite{pmlr-v70-kim17a}, CyCADA \cite{hoffman2018cycada} and \cite{pargan}. These methods regularize the GAN objective with Cycle-consistency by training an auxiliary GAN that performs the inverse mapping from the target domain to the source one. 

While CycleGAN \cite{zhu2017unpaired} is agnostic to the downstream task and operates entirely on the image-level, the mapping of CyCADA \cite{hoffman2018cycada} is learned in an online fashion together with the downstream task thus bringing both the image-level and the task-specific features close. While this makes larger domain shifts possible, it also makes the domain adaptation tied to a specific downstream task and the features used in it. We decided to go with the downstream task agnostic mehtod, first, because the domain shift between the synthetic and real pedestrians is relatively small, and second, because we want to allow any existing approach to leverage our DA method by simply training with the domain-adapted data as opposed to downstream-task specific approaches that require various modification to the training procedure. 

Therefore, we settle for ParGAN \cite{pargan}, which is based on CycleGAN, but different in the sense that it conditions the GANs on a parameter and then learns the parameterized mapping from the target domain the the union of the target and source domain equipped with the parameter. This work is an extension of \cite{zhu2017unpaired} that we use for the domain translation with several adjustments to the architecture.

\section{Model}
In this section we will describe our model for synthetic-to-real domain adaptation that can be applied to various geometry-based tasks. We assume that a labeled synthetic dataset is given. The images in the source and target domains are unpaired since we do not have examples of real-world images mapped to the synthetic representations. Our goal is to train a detection model on the synthetic data that would generalize to the real dataset of interest.  In our approach, we decouple the process of domain adaptation from the detection task and perform domain adaptation with the help of a ParGAN. The core of this domain adaptation model is a GAN \cite{goodfellow2020generative} consisting of a generator and a discriminator. The primary goal of the domain adaptation generator is to make it impossible for a discriminator to distinguish between the GAN-modified synthetic images and the target domain. However, we also want the image contents not to change too much, otherwise the labels would also change. Note that the method assumes that the differences between the synthetic and real domains is  mostly low-level, meaning that the coloring, contrast or lighting is different, but the high-level features, such as object shapes and locations, the overall geometry is similar.

\begin{figure}[ht!]
\centering
\includegraphics[width=0.235\textwidth]{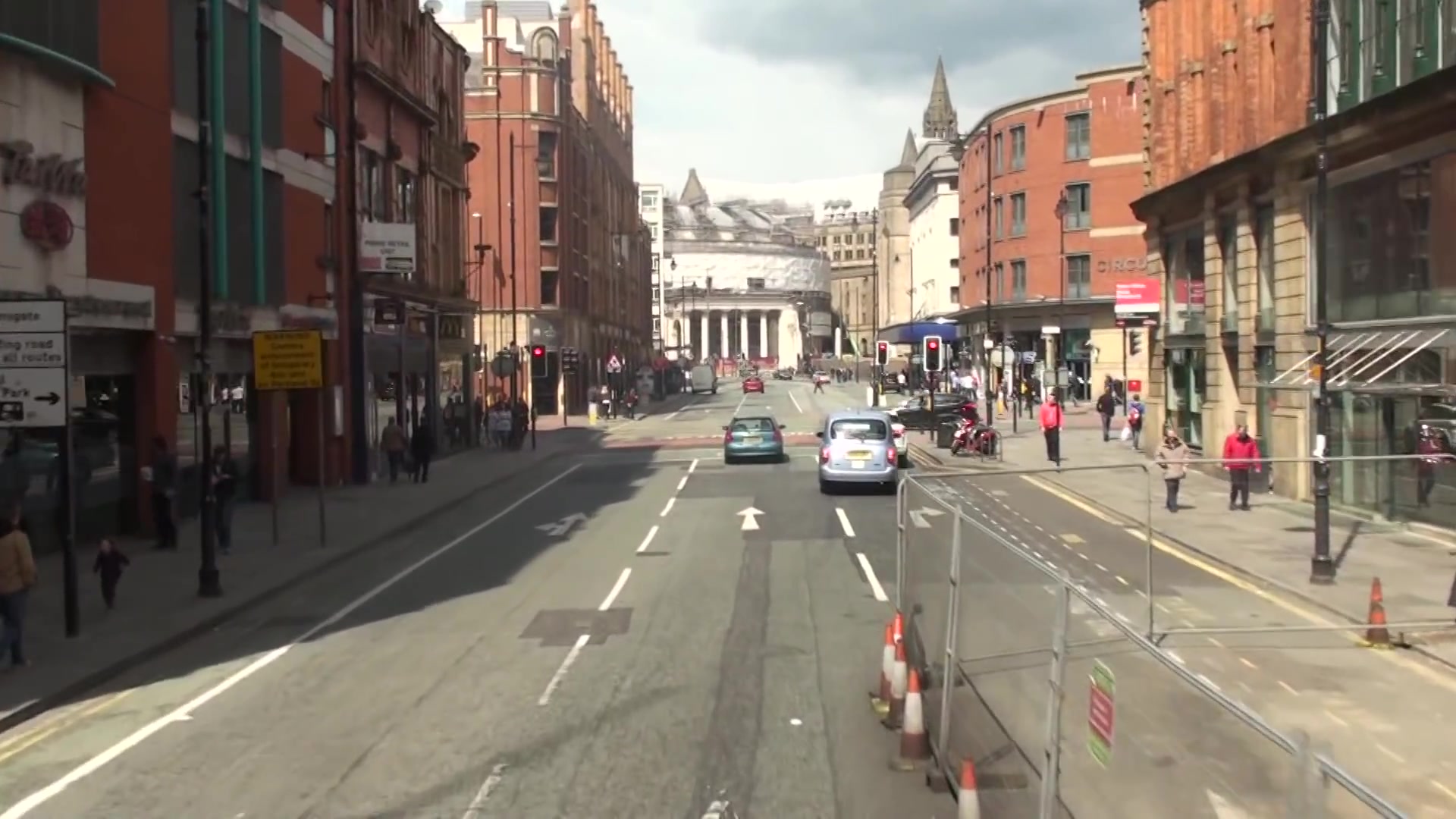}
\includegraphics[width=0.235\textwidth]{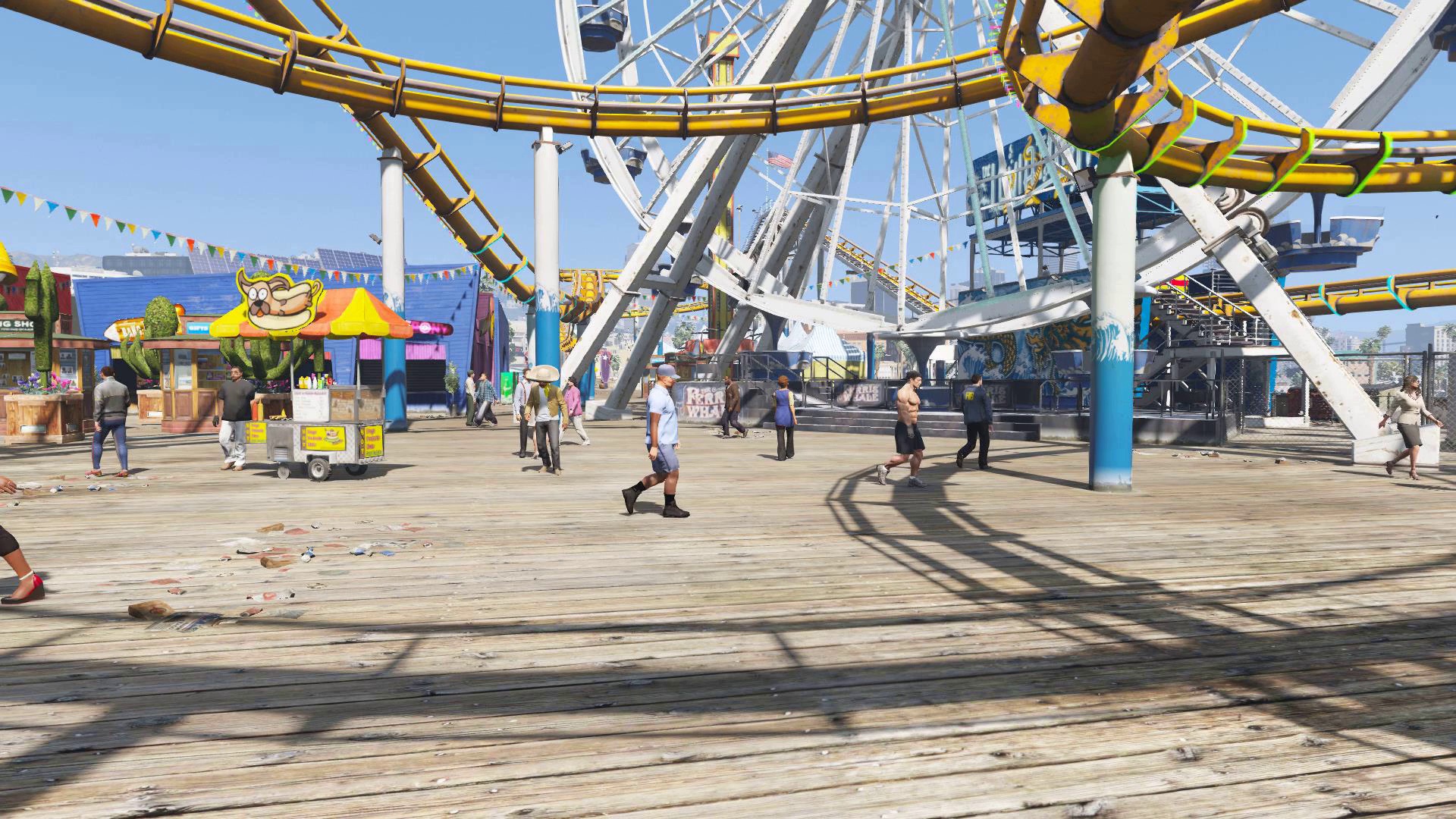}
\caption{2 frames from MOT17 (left) and MOTSynth (right) that represent the domain closeness}
\label{fig:synth_demo}
\end{figure}

In particular, let $\mathbf{X}^s = \{X_i^s, y_i^s\}_{i=1}^{N^s}$ be the synthetic (source) dataset of images $X^s_i\sim p_{X^s}$  and their labels (bounding boxes) $y_i^s\sim p_{y\mid X^s}$ and $\mathbf{X}^t = \{X_i^t\}_{i=1}^{N^t}\sim p_{X^t}$ be the dataset of real (target domain) images. We will also use $X^{s,t}$ to define an image that is equally likely to be from the source or from the target domain $p_X(\cdot) = (p_{X^s}(\cdot) + p_{X^t}(\cdot))/2.$ We want to construct a map $G(x^s),$ such that $p_{G(X^s)}\approx p_{X^t}.$
However, when constructing the generative model, we want to take into account the high quality of the synthetic images. Consider the two images on figure \ref{fig:synth_demo}: the left one is from MOTSynth dataset and the right one is a frame from a video taken by a bus on a busy intersection in MOT17 dataset. 

The two look very similar both on the high level (shapes and figures) and on the low level (color and lighting) in the sense that it is hard to tell which one is real and which one is synthetic. We thus turn to differentiating the images not in terms of their origin, whether they are synthetic or real, but based on their embedding features. This brings the domain adaptation task to translating from the synthetic domain to the domain of images with high "realness", be they synthetic or real. This not only makes our problem better-posed, but also reduces the GAN confusion as shown in \cite{mirza2014conditional}. The domain adaptation problem then falls into the realm of conditional unpaired image-to-image translation, which can be solved by a Parametic GAN \cite{pargan}.

Let $p(X)\in\mathbb{R}$ be a parameter that specifies how close the image is to the target dataset. Our domain adaptation model trains a generator $G(x^s,p)$ that maps a pair of a synthetic image and a parameter value $p$ to an image  $X^{s,t}$ such that $p(G(x^s,p))\approx p.$ For real images  $p(x^t)\gtrapprox1$  and for synthetic images $p(x^s)\lessapprox0.$  For generation we choose $p=1,$ the generated dataset $\{(G(x^s_i, 1), y^s_i)\}_{i=1}^{N^s}$ then has a smaller domain gap with the real domain than the original synthetic one and leads to better generalization of the detection model trained on it. Note that if we choose $p(X^s) = 0, p(X^t)=1,$ the ParGAN will turn into a CycleGAN \cite{zhu2017unpaired}. 

\subsection{Parameterizing the gap}
\begin{figure}
    \centering
    \includegraphics[width=\linewidth]{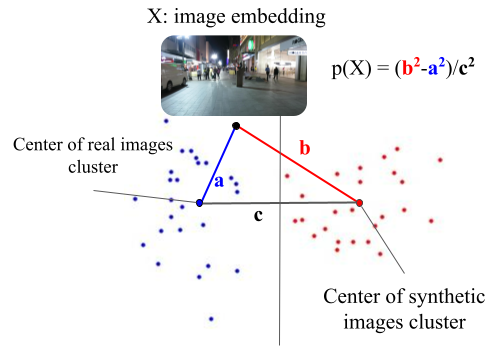}
    \caption{Calculating the realness of image $X$ in the embedding space, the "realness parameter is proportional to the distance to the hyperplane, separating the centers of real (blue) and synthetic (red) image embeddings}
    \label{fig:parameter}
\end{figure}

While there are countless ways to define the "realness" of the image, for our purposes we defined it through the normalized difference between the squared distance to the center of the synthetic dataset  $\bar X^s = \frac1{N^s}\sum_{i=1}^{N^s}f\left(X_i^s\right)$ and the squared distance to the real dataset $\bar X^t = \frac1{N^t}\sum_{i=1}^{N^t}f\left(X_i^t\right),$ both taken in the embedding space, where $f(\cdot)$ is the embedding function: 
\begin{equation}
    p(X) = \frac{\left\|f(X) - \bar X^s\right\|^2 - \left\|f(X) - \bar X^t\right\|^2}{\left\|\bar X^t - \bar X^s\right\|^2}.\label{eq:parameter}
\end{equation}
If $X$ is embedded onto the center of the synthetic dataset $\bar X^s$ then $p(X)=-1,$ and if it is embedded onto the center of the real dataset $\bar X^t$ then $p(X) = 1.$ An example of such calculation is on Fig. \ref{fig:parameter}

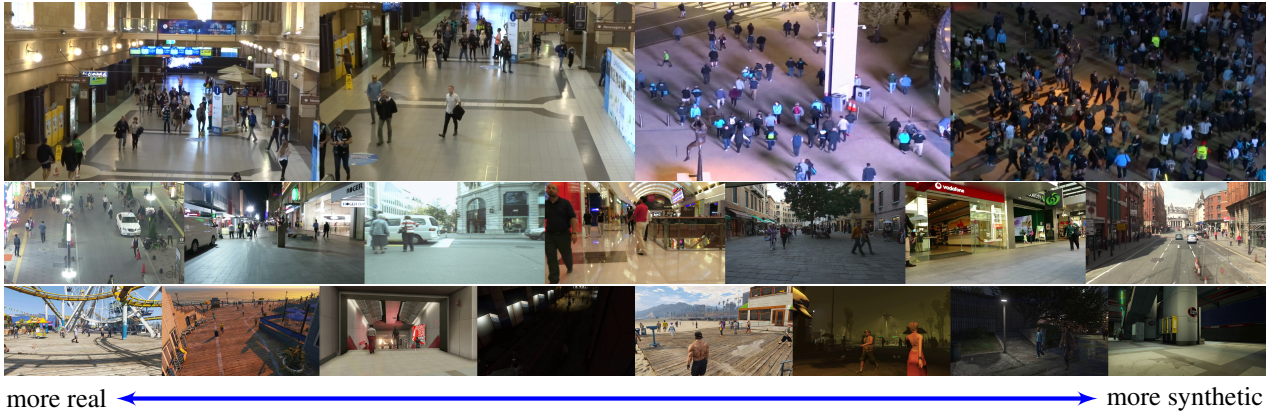
\begin{figure*}[ht!]
    \centering
    \includegraphics[width=0.96\textwidth]{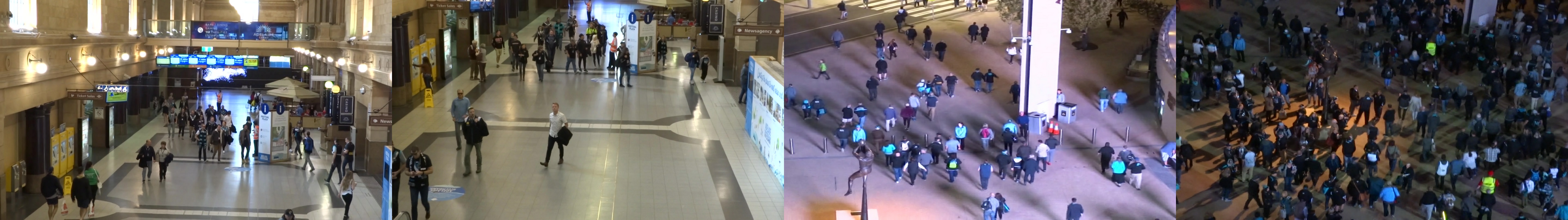}
    \includegraphics[width=0.96\textwidth]{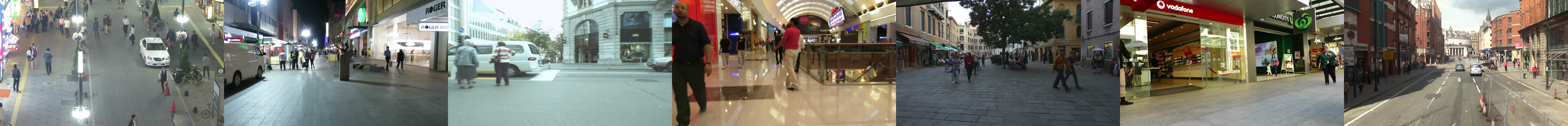}
    \includegraphics[width=0.96\textwidth]{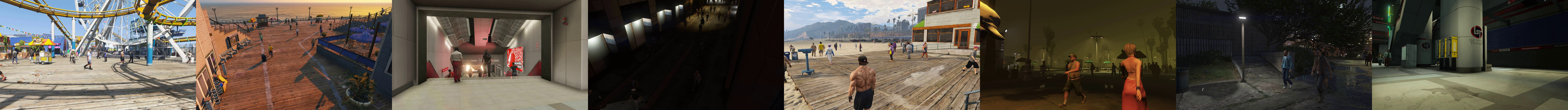}
    \begin{tikzpicture}
    \node (A) at (0, 0) {more real};
    \node (B) at (15, 0) {more synthetic};
    \draw [ultra thick, blue, latex' -latex'] (A) -- (B);
    \end{tikzpicture}
    \caption{Frames from MOT20 (top), MOT17(center), MOTSynth(bottom) in the decreasing order of the parameter}
    \label{fig:videos_ordered_by_p}
\end{figure*}

For at least half of the real dataset $p(X)\geq1$ and for at least half of the synthetic dataset $p(X)\leq-1.$ This parameter acts as a classifier between the real and synthetic data while also being fast to compute and parallelizable, it is also representative of the distance to the target dataset if the distance to the synthetic dataset is fixed, which is approximately the case for our model since we penalize large changes as discussed in the subsection \ref{subsec:pargan}. To test if our choice is meaningful, we plotted the images from the MOT17, MOT20 and MOTSynth datasets in order of the decreasing parameter on \cref{fig:videos_ordered_by_p} and observed that indeed the images with a higher parameter values look more realistic than the ones with the lower parameter.

\subsection{ParGAN Engineering}\label{subsec:pargan}

Our ParGAN model has to address several challenges, namely it has to lessen the domain gap: $p_{G(p_{X^s},p)}\approx p_{X\mid p(X)=p},$ while also preserving the content of the image: $G(p_{X^s},p)\approx X^s,$ otherwise the labels would have to be changed, so we made some adjustments to the loss function from \cite{pargan}. 

{\bf Architecture.} The ParGAN architecture we used 
%is shown on \textbf{figure} and 
is similar to the one used in \cite{zhu2017unpaired,pargan}. It consists of two GAN models: the "forward" one, which samples from the target distribution and has generator $G(X, p)$ and discriminator $D(X, p),$ and the "inverse" GAN model with generator $G_\inv(X, p)$ and discriminator $D_\inv(X, p).$ 
The Generator architectures are adapted from \cite{johnson2016perceptual}, this network has an encoder-decoder type architecture and contains three convolutions, 6 residual blocks, fractionally-strided convolutions, and one convolution that maps features to RGB. The discriminator networks are $70\times70$  PatchGANs \cite{isola2017image}, which classify $70\times70$ overlapping image patches  into real and fake while having a smaller number of parameters than standard GAN discriminators. However, we changed the parGAN \cite{pargan} model in that we use resize-convolutions for upsampling as suggested here \cite{aitken2017checkerboard} to avoid checkerboard patterns. 
%The resize-convolutions helped us get rid of the checkerboard pattern, while

{\bf Loss function.} The loss is a sum of the three functions that ensure the optimal generators satisfy the following:
\begin{itemize}
    \item $\mathcal{L}_\GAN:\;p_{G(X^s, p)}(\cdot)\approx p_{X\mid p(X)=p}(\cdot)$ -- the "forward" generator transforms a synthetic image to an image with the specified parameter value (the distribution of the generator output is $X\mid p(X) = p$)
    \item $\mathcal{L}_{GAN,inv}:\;p_{G_\inv(X, p(X))}(\cdot)\approx p_{X^s}(\cdot)$ -- the "inverse" generator transforms a real or synthetic image with parameter $p$ into a synthetic image (the distribution of the "inverse" generator output is $p_{X^s}(\cdot).$)
    \item $\mathcal{L}_\cyc:$ the "forward" and "inverse" generators are inverse of each other conditioned on the parameter value: $G_\inv(G(X, p), p)\approx G(G_\inv(X, p), p)\approx X$
\end{itemize}
The way $\approx$ is defined specifies the loss and consequently the end model parameters. In \cite{zhu2017unpaired}, the Jensen-Channon divergence \cite{goodfellow2020generative} (the cross-entropy loss) is used to measure distance $(\approx)$ in  $\mathcal{L}_\GAN,\mathcal{L}_{\GAN,\inv},$ while the authors of \cite{pargan} change it to the least-squares loss \cite{lsgan}. Our preliminary experiments showed that Wasserstein loss with gradient penalty \cite{gulrajani2017improved} led to both better quality images and more stable training, so we chose it for our ParGAN model, while the cycle-consistency loss $\mathcal{L}_\cyc$ was left to be the mean $\ell_1$ distance. Finally, we also added the identity loss, which the authors of \cite{zhu2017unpaired} used for the paintings $\leftrightarrow$ photo translation to avoid the day/night inversion, which resulted in the following total loss
\begin{align}
    \mathcal{L}(G, &G_\inv, D, D_\inv)=\mathcal{L}_\GAN(G, D) \label{eq:total_loss}\\
    &+ \mathcal{L}_\GAN'(G_\inv, D_\inv)+ \lambda_\cyc \mathcal{L}_\cyc(G, G_\inv)\nonumber\\
    &+ \lambda_{id} \left(\mathcal{L}_{id}(G) + \mathcal{L}_\Id'(G_\inv)\right),\nonumber
\end{align}
where the individual components are
\begin{align}
    \mathcal{L}&{}_\GAN(G, D)\label{eq:GAN1}\\
    &=\mathbb{E}\left[D(G(X^s, p(X)), p(X))]-D(X, p(X))\right] \nonumber\\
    \mathcal{L}&{}_\GAN'(G_\inv, D_\inv)\label{eq:GAN2}\\
    &=\mathbb{E}\left[D_\inv(G_\inv(X, p(X)), p(X))]-D_\inv(X^s, p(X))\right] \nonumber\\
    \mathcal{L}&{}_\cyc(G, G_\inv)\\
    &=\mathbb{E}\left[\left\|G_\inv(G(X^s, p(X)), p(X)) - X^s\right\|_1\right]/2\nonumber\\
    &\quad+\mathbb{E}\left[\left\|G(G_\inv(X, p(X)), p(X)) - X\right\|_1\right]/2\nonumber\\
    \mathcal{L}&{}_{id}(G)=\mathbb{E}\left[\left\|G(X^s, p(X)) - X^s\right\|_1\right]\\
    \mathcal{L}&{}_{id}'(G_\inv)=\mathbb{E}\left[\left\|G_\inv(X, p(X)) - X\right\|_1\right].
\end{align}
The loss is minimized over the parameters of the generators $G,G_\inv$ and maximized over the parameters of the discriminators $D,D_\inv,$ which are additionally regularized with gradient penalty to enforce approximate 1-Lipschitzness. 

\subsection{Detection Model}
We used a one-stage RetinaNet detection model \cite{lin2017focal} with a ResNet-50 backbone and a 256-dimensional FPN \cite{lin2017feature} head. We chose the RetinaNet detector since it achieves performance similar to Faster RCNN, while being a one-stage detector, thus simplifying the training process.

\section{Training Setup}
In this section we briefly describe the training process and parameters we used in our experimental evaluations.
\subsection{ParGAN training}
The training scheme is shown in \cref{fig:training_scheme}. We trained the ParGAN model on full-resolution MOTSynth $\leftrightarrow$ respective dataset (MOT17, MOT20 or CityPersons) domain adaptation task. We used a batch size of 1 and randomly cropped patches of size $128\times128$ from the images similarly to CycleGAN and ParGAN.

The parameter calculation was different for different datasets. For MOT17, the embedding for parameter calculation \eqref{eq:parameter} was the 127-dimensional image embedding trained internally (the top-results of \url{http://www.kaggle.com/competitions/google-universal-image-embedding}, perform very similarly to the embedding). We later discovered that ImageNet-trained classification features perform equally well, so for MOT20, we chose the 2048-dimensional features extracted from Imagenet-trained ResNet-50. We also filtered out the MOTSynth images at night time to bring the two domains closer, since the images in the target domains are not as dark. The inference and subsequent detection model training were however performed on the whole MOTSynth dataset. The ParGAN models were trained from scratch on GPU Tesla V100 for 100000 steps with cycle consistency weight $\lambda_\cyc=10$ and identity loss weight $\lambda_\Id=1.25.$

% \textbf{Ablation Study}
% To choose the right loss/architecture parameters we performed an ablation study on MOT17. We downsized the MOTSynth dataset frames to $480\times720$ resolution and used an MNASnet (\textbf{figure out the architecture}) for faster turnaround. We evaluated the results on MOT17 train split. Since we did not use the MOT17 labels in the training process, we think the results might be representative of the true performance on MOT17 test split, however there might be a discrepancy since the MOT17 train data was seen by the ParGAN. We used mean average precision (mAP) at IoU threshold of 0.5 to rank the detection performance.

% \textbf{Architecture Choices}

% \textbf{Loss Function Choices}

\subsection{Detection model training}

For the detection model, the imagery generated by the ParGAN model was combined with the original MOTSynth labels, and then used to train for 250000 steps and a batch size of 1024 distributed across 64 chips. We randomly scaled the images to $[0.3,2.0]\times$ original size and cropped them back  to augment.

\subsection{Ablation Studies}
In this subsection, we present evidence to support our choice of ParGAN as the domain adaptation model and its loss function. Since training the domain adaptation and detection models is time and resource-consuming, we chose a smaller detection model -- a 10-layer tunable MNASNET \cite{tan2019mnasnet} as the backbone for detection in the ablation studies. Additionally, we resized the images to 480p for both synthetic and real domains, the rest of the training parameters were unchanged. We did a hyperparameter search to determine the learning rate and loss function weights. We trained the domain adaptation model on the train portion of the MOT17 dataset without using any real labels and calculated the mAP based on the real labels of the train part of MOT17 (note that labels of the test split of MOT17 are not public). We used the final mAP results \ref{tab:ablation} to choose the final model for domain adaptation. We used mAP of a detector trained purely on the synthetic data as a baseline to compare. 
% In the latter 2 rows of \ref{tab:ablation} "Wasserstein loss" indicate the change of the GAN loss functions $\mathcal{L}_{GAN},\mathcal{L}_{GAN,inv},$ from the Jensen Shannon divergence to the 1-Wasserstein distance as in  \eqref{eq:GAN1} and \eqref{eq:GAN2}.
for the experiment in the fourth row, we set the parameter to be a constant for all the images, thus getting a cycle GAN with identity loss and Wasserstein distance instead of the Jensen-Shannon divergence used in $\mathcal{L}_{GAN}$ and $\mathcal{L}_{GAN, inv}$.

\begin{figure}[ht!]
    \centering
    \begin{tabular}{|c|c|}
    \hline
    Method & MNASNET AP\\
    \hline
    no domain adaptation & 0.6717\\
    CycleGAN & 0.7039 \\
    CycleGAN with identity loss & 0.7200 \\
    ParGANDA \eqref{eq:total_loss}, no parameter& 0.7292 \\
    ParGANDA \eqref{eq:total_loss} & 0.7346 \\
    \hline
    \end{tabular}
    \captionof{table}{MOT17 dataset ablation study with MNASNET backbone}
\label{tab:ablation}
\end{figure}

% \vspace{-1em}
% \begin{table}[h]
% \centering
% \caption{Ablation study.
% \vspace{-1em}}
% \label{tab:ablation}
%     \resizebox{0.475\textwidth}{!}{
%     \large
%     \setlength\tabcolsep{1pt}
%     \begin{tabular}{*{10}{c}}
%         \toprule
%        Method & MNASNET AP\\
%         \midrule
%         no domain adaptation & 0.6717\\
%         CycleGAN & 0.7039 \\
%         CycleGAN with identity loss & 0.7200 \\
%         CycleGAN with identity loss and Wasserstein GAN & 0.7292 \\
%         ParGAN with identity loss and Wasserstein GAN & 0.7346 \\
%         \bottomrule
%     \end{tabular}
%     }
%     \vspace{-1em}
% \end{table}
The addition of the identity loss and the change of GAN loss to Wasserstein loss increased the AP by 0.03 total and the addition of the parameter to the domain adaptation increased the AP by 0.005, which we believe justifies the change in the loss function and the use of ParGAN.

\section{Evaluation}
The goal of many domain adaptation tasks in computer vision is to produce images that are visually appealing, high-quality and free of artifacts. However, in our case the downstream task metric is what ranks the domain adaptation models, thus we are not particularly interested in the visual quality of the images. We also note that for some downstream tasks the artifacts of a generative model can serve as augmentations. With that being said, we presented several examples of the ParGAN outputs on \cref{fig:inference}. 
To draw conclusions about the performance of the domain adaptation method, we evaluate it on a pedestrian detection task by calculating the mean average precision for intersection
over union (IoU) threshold equal to 0.5 and averaged over recall values of 0.1-1.0.

\textbf{MOT Challenge}
\begin{figure}[ht!]
    \centering
    \begin{tabular}{|c|c|c|}
    \hline
    Eval&Train& AP\\
    \hline
    \multirow{2}{*}{MOT17}&MOTSynth&0.79\\
    &MOTSynth+ParGANDA&0.85\\
    \hline
    \multirow{2}{*}{MOT20}&MOTSynth\cite{fabbri2021motsynth}&0.62\\
    &MOTSynth+ParGANDA&0.69\\
    \hline
    \end{tabular}
    \captionof{table}{MOT challenge evaluation results, no labels from the real datasets were used for training, ParGANDA=domain adaptation of MOTSynth with ParGANDA}
\label{tab:mot_results_no_labels}
\end{figure}

\setlength{\tabcolsep}{4pt}
\begin{figure}[ht!]
    \centering
    \begin{tabular}{|c|c|c|c|c|c|}
    \hline
    \multirow{2}{*}{Eval}&\multirow{2}{*}{Method}&MOT&real&temp.&\multirow{2}{*}{AP}\\
    &&Synth&labels&info&\\\hline
    \multirow{7}{*}{MOT17}&FRCNN\cite{fabbri2021motsynth}&\cmark&\xmark&\xmark&0.80\\
    &ParGANDA&\cmark&\xmark&\xmark&\textbf{0.85}\\
    \cline{2-6}
    &FRCNN\cite{fabbri2021motsynth}&\cmark&\cmark&\xmark&0.80\\
    &FRCNN\cite{ren2015faster}&\xmark&\cmark&\xmark&0.72\\
    &ZIZOM\cite{lin2018graininess}&\xmark&\cmark&\xmark&0.81\\
    &SDP\cite{yang2016exploit}&\xmark&\cmark&\xmark&0.81\\
    &SGT\_det\cite{hyun2022detection}&\xmark&\cmark&\cmark&\textbf{0.90}\\
    \hline\hline
    \multirow{4}{*}{MOT20}&FRCNN\cite{fabbri2021motsynth}&\cmark&\xmark&\xmark&0.62\\
    &ParGANDA&\cmark&\xmark&\xmark&\textbf{0.69}\\
    \cline{2-6}
    &FRCNN\cite{fabbri2021motsynth}&\cmark&\cmark&\xmark&0.72\\
    &GNN\_SDT\cite{wang2021joint}&\xmark&\cmark&\cmark&\textbf{0.81}\\
    \hline
    \end{tabular}
    \captionof{table}{Comparison of ParGANDA with top-performing MOT17/MOT20 detection models with the distinction based on whether they use synthetic and/or real labels and temporal data}
\label{tab:mot_results}
\end{figure}

We evaluate our final detection models on the MOT17Det and MOT20Det detection datasets \cite{milan2016mot16} on the private server of the challenge. The results we got are in \cref{tab:mot_results_no_labels}. They indicate that indeed, ParGANDA leads to an increase in performance of the detection model, $0.06$ mAP for MOT17 and $0.07$ mAP for MOT20. In \cref{tab:mot_results_no_labels} we compare the results of ParGANDA  together with the top-ranking models of the challenge (we only included models for which we could find citations). Even more optimistic results come from the comparison of ParGANDA with models that have access to real labels: for MOT17 ParGANDA outperforms all the methods trained on real data that do not make use of temporal information coming from the videos of MO17.

% ToDO: say something about the results on MOT20
% \setlength\tabcolsep{5pt}

\textbf{CityPersons Dataset}

\begin{figure}[ht!]
    \centering
    \begin{tabular}{|c|c|c|}
    \hline
    Eval&Train& AP\\
    \hline
    \multirow{2}{*}{CityPersons}&MOTSynth&0.54\\
    &MOTSynth+DA&0.66\\
    % \hline
    % \multirow{2}{*}{CalTech Pedestrian}&MOTSynth&0.457\\
    % &MOTSynth+DA&0.462\\
    \hline
    \end{tabular}
    \captionof{table}{CityPersons evaluation results}
\label{tab:small_dataset_results}
\end{figure}
We also trained ParGAN for domain adaptation on the CityPersons dataset, the results are in \cref{tab:small_dataset_results}. As in the previous experiments, we are most interested in the difference between the detection model trained purely on the synthetic data versus the detection model trained on the ParGAN-translated synthetic data, and we observed a $0.12$ increase in mAP.

\subsection{Qualitative results}
To qualitatively analyse the domain adaptation performance, we begin with visual inspection of the generated images. Some of the GAN outputs can be found on \cref{fig:inference}. Overall, the image quality is good and one can clearly see that the low-level features (lighting, coloring) of the adapted images are mimicking the ones of the target dataset better than the synthetic ones.

\begin{figure}[ht!]
    \centering
    \includegraphics[width=\linewidth]{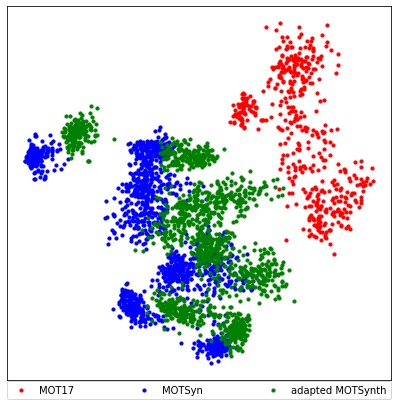}
    \caption{2-dimensional PCA of the image embeddings for 8 MOTSynth videos sampled at 1/10 framerate and MOT17 dataset}
    \label{fig:pca_mot17}
    %TODO: fix the legend to adapted motsynth
\end{figure}

We further aim to evaluate how well the ParGAN performs the direct task assigned to it: making the parameter of synthetic images \eqref{eq:parameter} close to the one of the real ones. To visualize the change on \cref{fig:pca_mot17} we plot a 2-dimensional PCA embedding of the target dataset MOT17 and the original MOTSynth dataset, and then add the representation of the ParGANDA adapted MOTSynth images. 
We only chose 8 MOTSynth videos for the plot to make the represented data balanced. Overall, we clearly see the movement of the MOTSynth dataset towards the target real dataset. However, the datasets are not indistinguishable on the plot. This probably indicates that the ParGAN loss and its architecture prevents it from large image changes. But this shortcoming is also an upside for the pedestrian detection task – we need the geometry to stay consistent after domain adaptation; drastic geometric changes would invalidate the accuracy of the known synthetic labels.

\section{Limitations and Discussion}
\begin{figure}
    \centering
    \begin{tabular}{cc}
    Synthetic image & ParGAN output\\
         \includegraphics[trim={9.5cm 0.5cm 5cm 5cm},clip,width=0.49\linewidth]{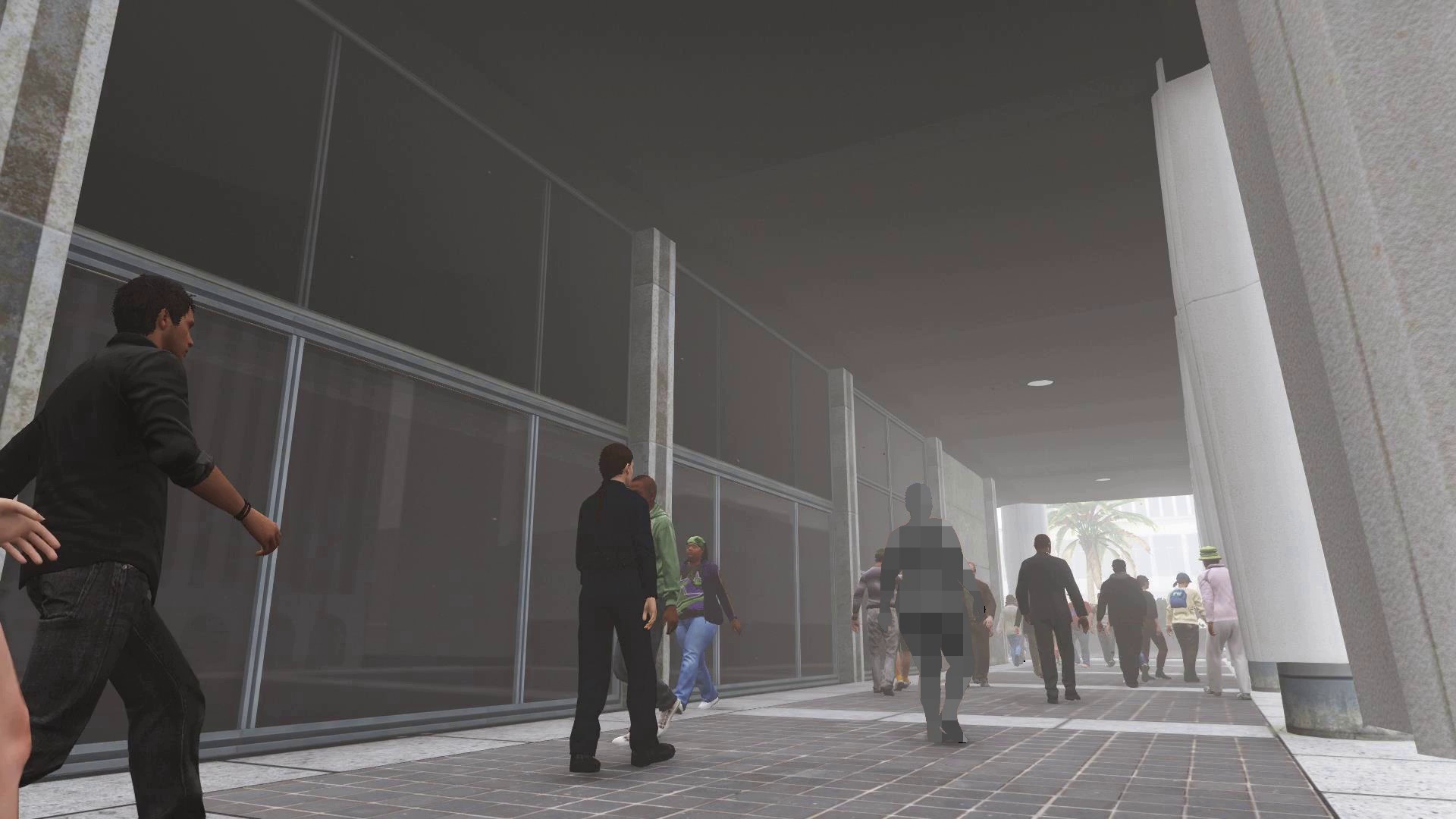}& \includegraphics[trim={9.5cm 0.5cm 5cm 5cm},clip,width=0.49\linewidth]{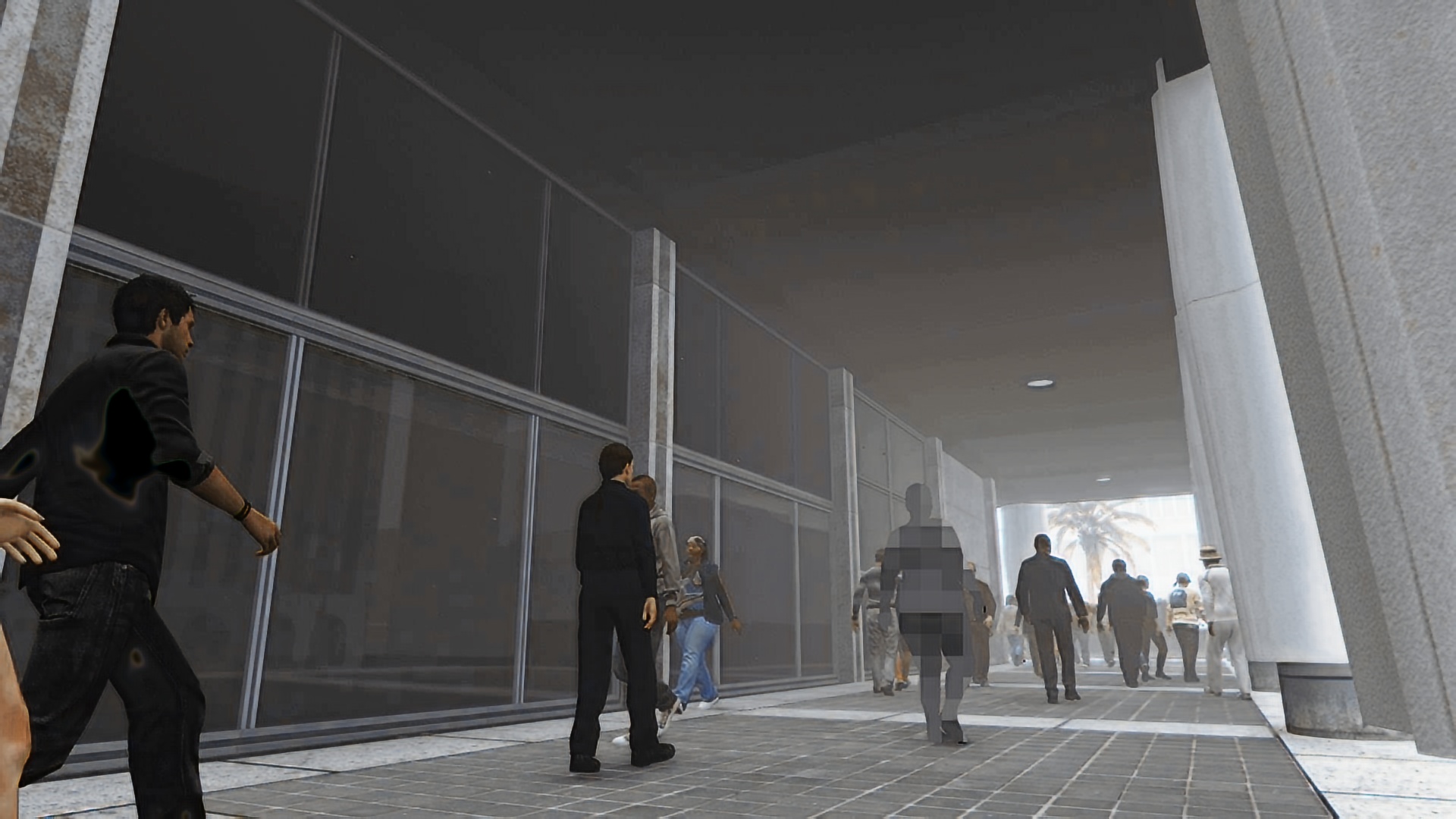}\\
         \includegraphics[trim={10cm 2cm 0cm 2cm},clip,width=0.49\linewidth]{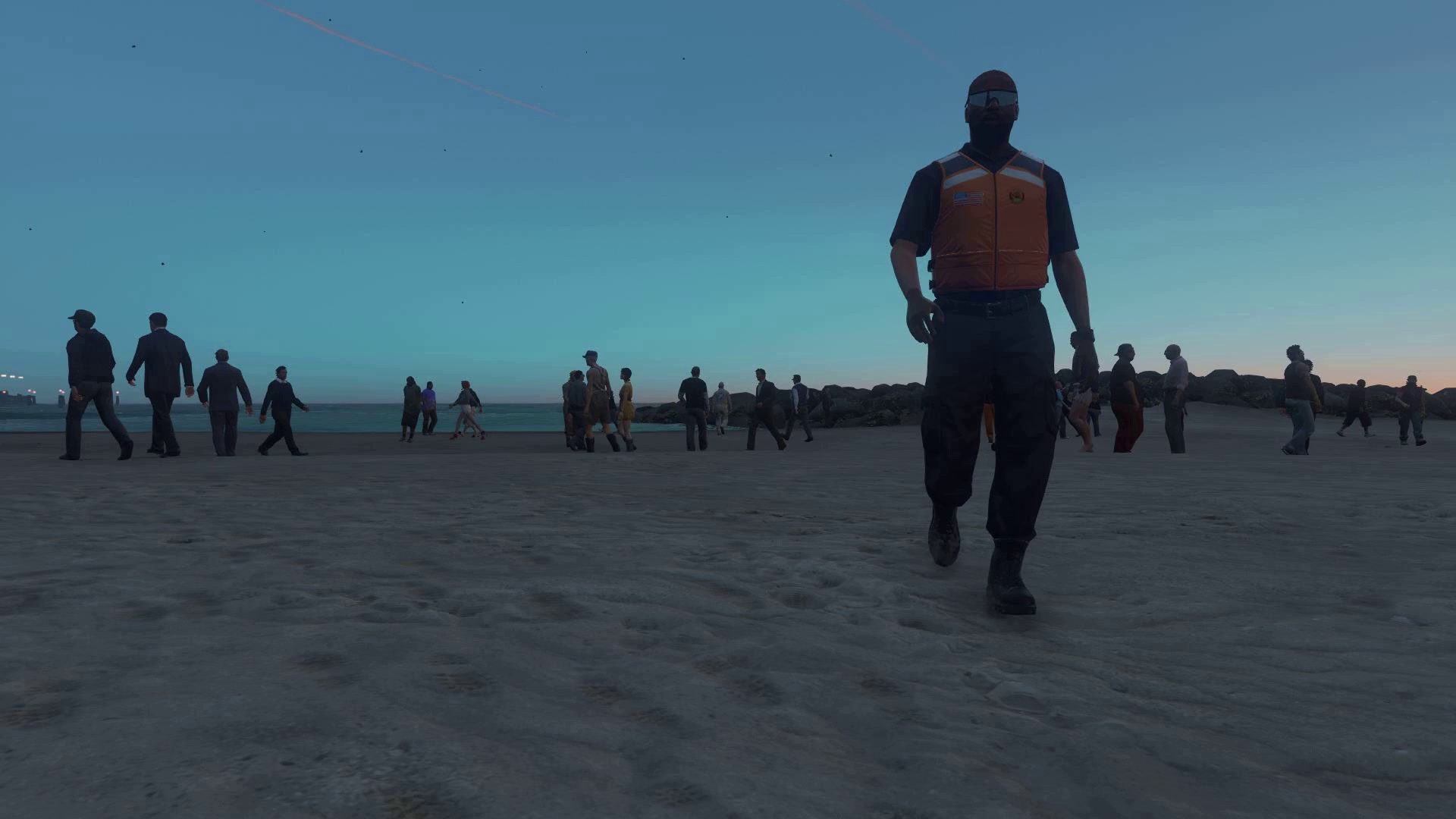}& \includegraphics[trim={10cm 2cm 0cm 2cm},clip,width=0.49\linewidth]{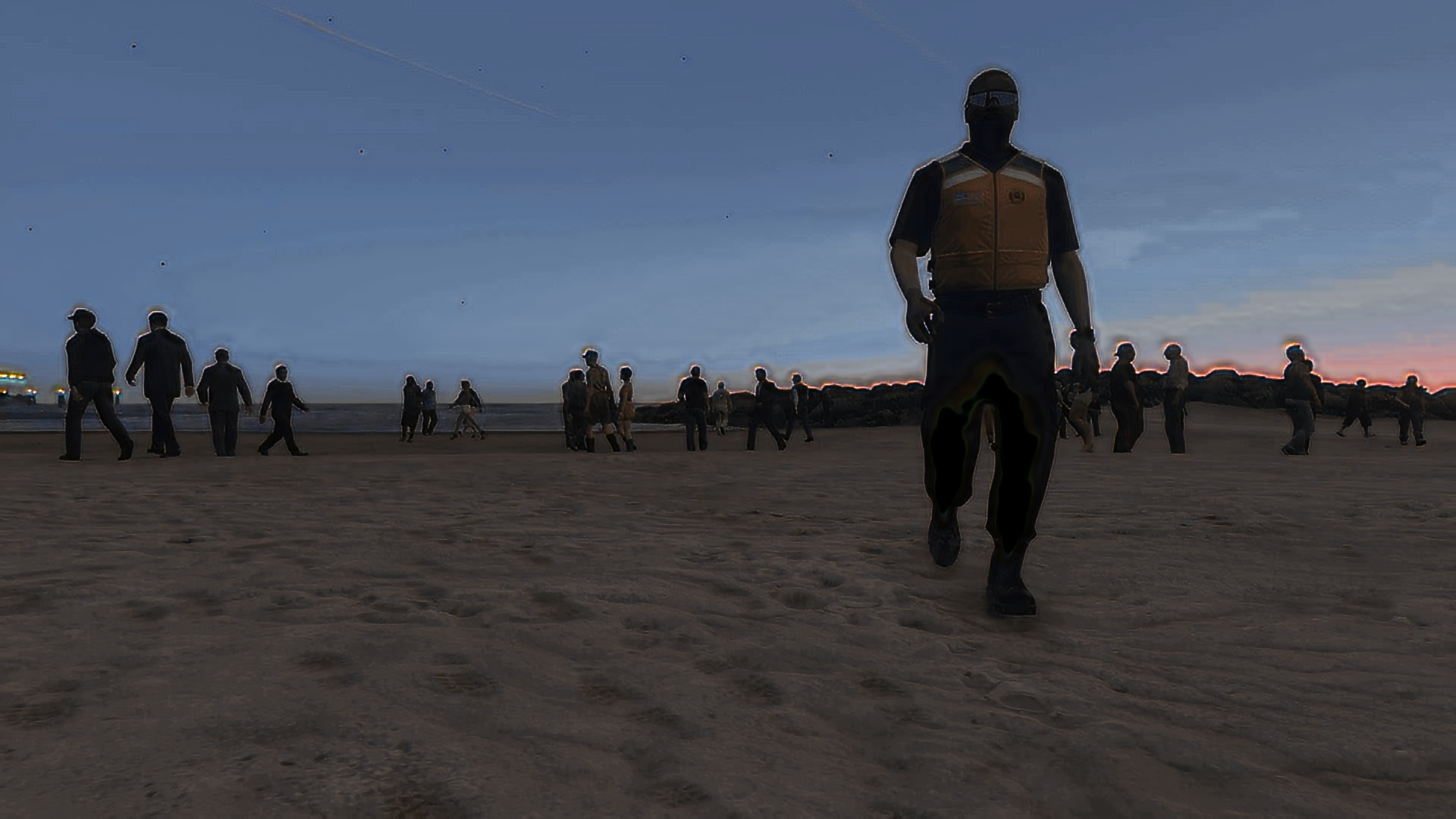}\\
         \includegraphics[trim={1cm 0cm 4cm 2cm},clip,width=0.49\linewidth]{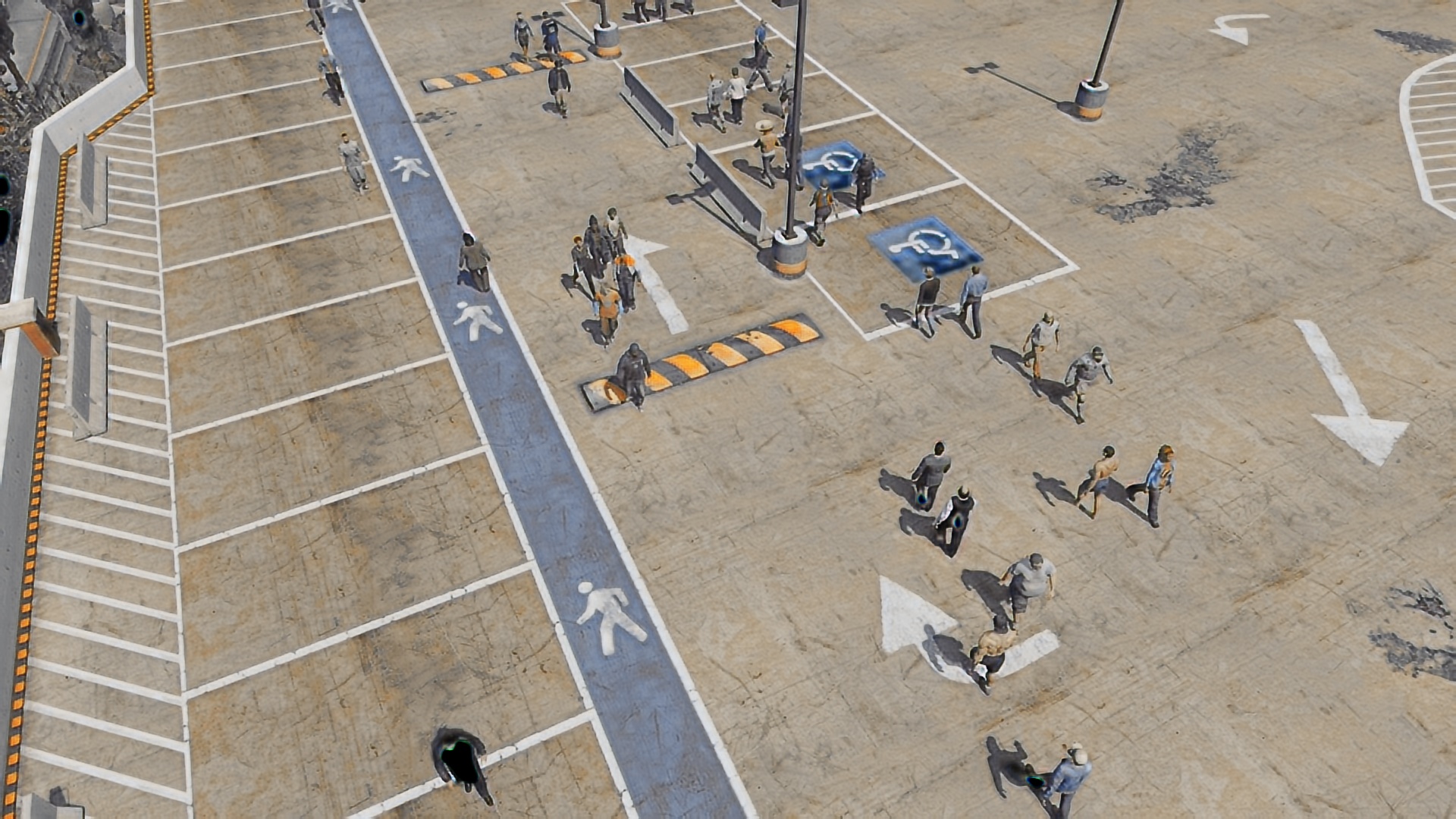}& 
         \includegraphics[trim={1cm 0cm 4cm 2cm},clip,width=0.49\linewidth]{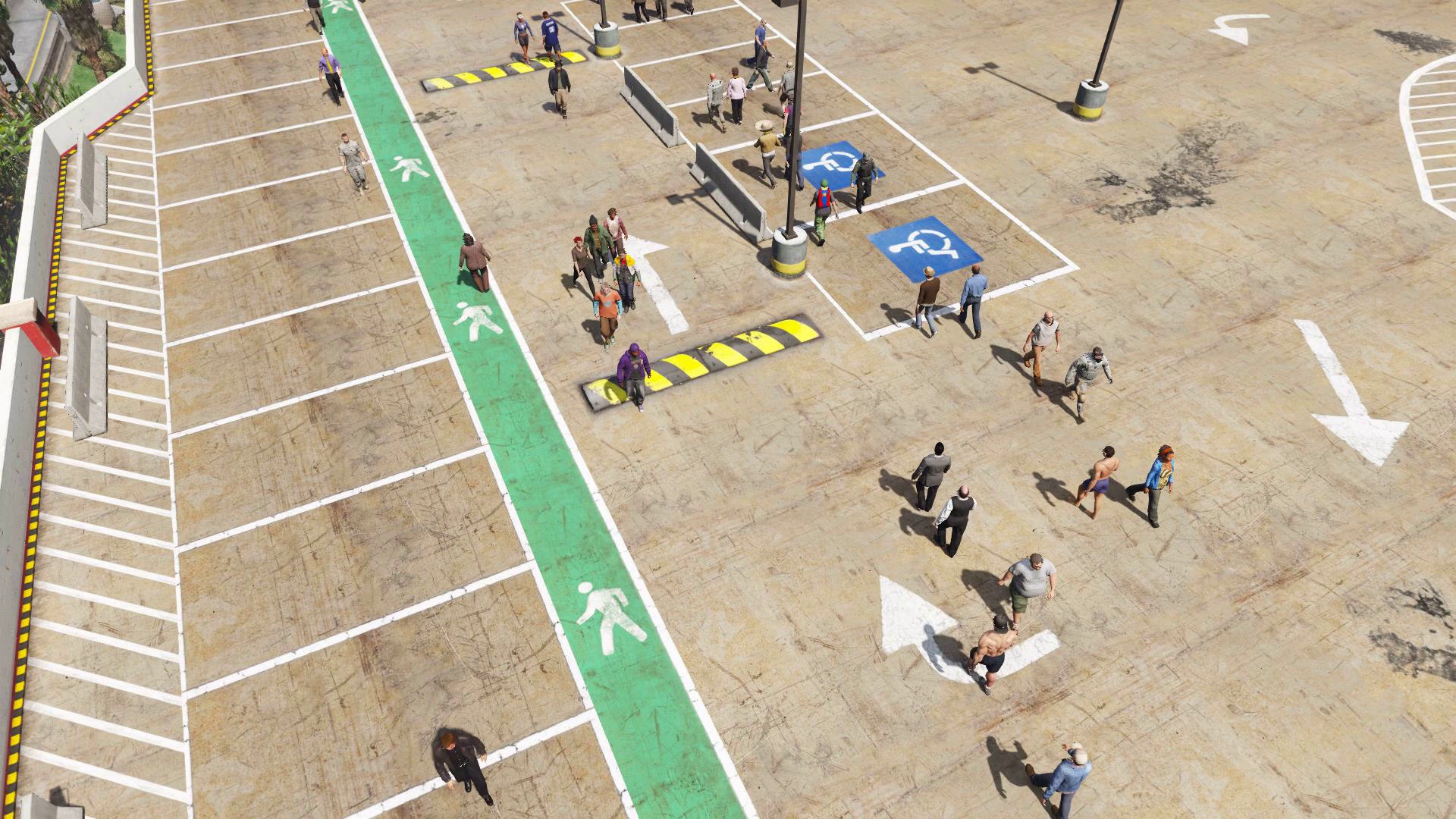}\\
         \includegraphics[trim={1cm 0cm 4cm 2cm},clip,width=0.49\linewidth]{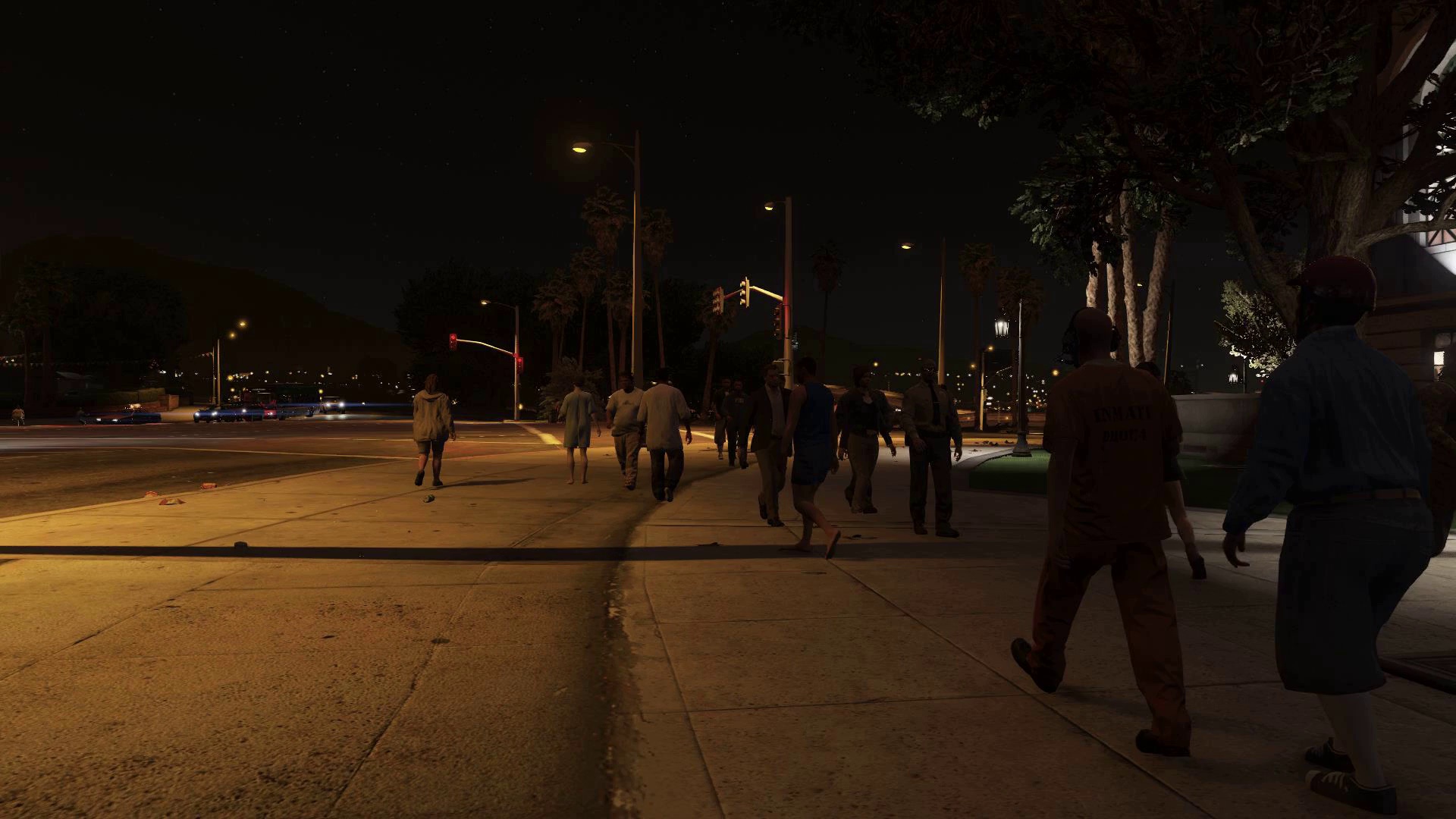}& 
         \includegraphics[trim={1cm 0cm 4cm 2cm},clip,width=0.49\linewidth]{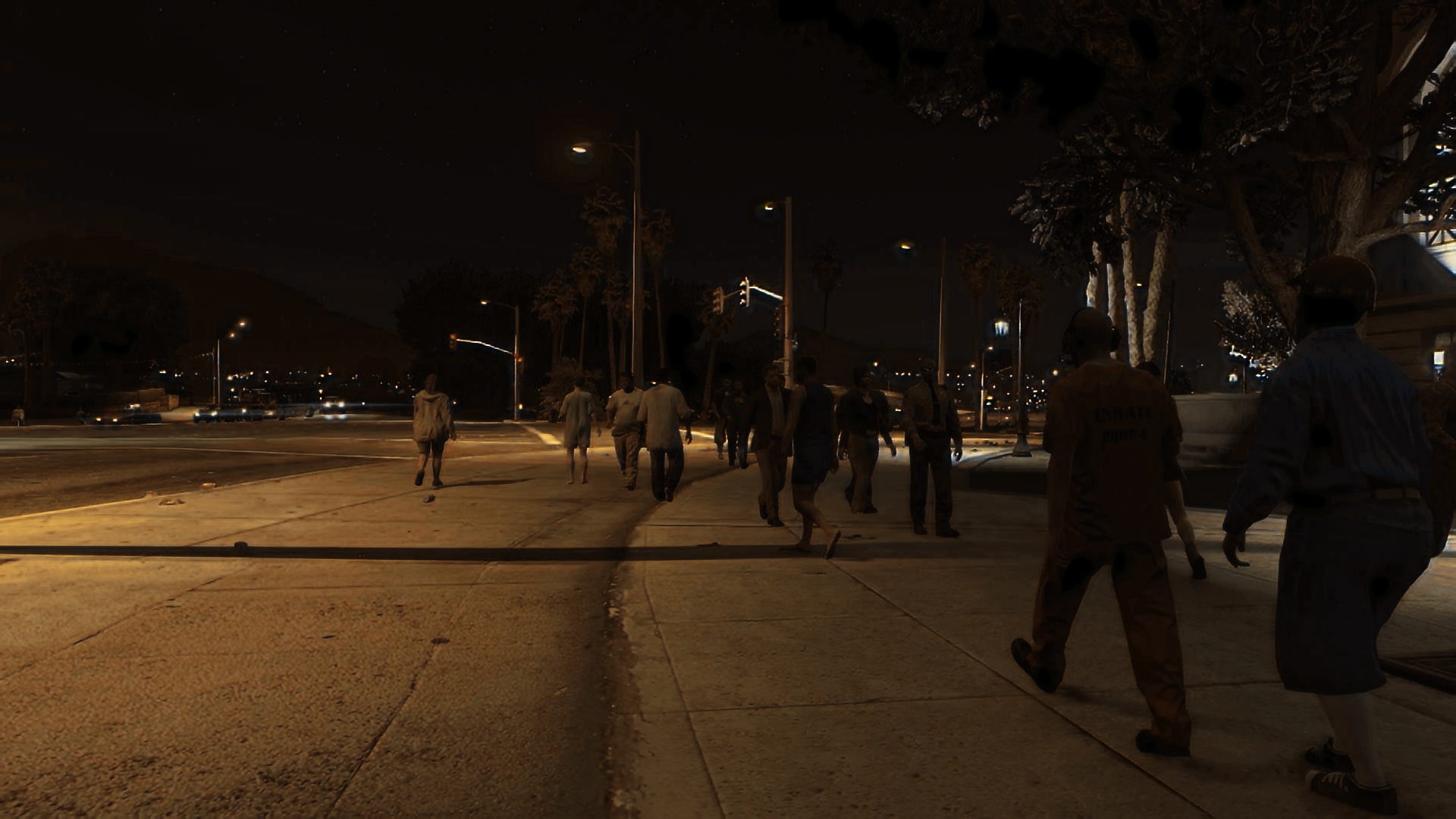}
    \end{tabular}
    \caption{Common failures of the ParGAN model}
    \label{fig:failures}
\end{figure}

Our experiments indicate that the ParGANDA can improve the precision of detection models and also produce visually appealing images. However, the quality of the generated images is not uniformly high. Fig. \ref{fig:failures} shows some of the typical failures of the domain adaptation model.

\setlength{\tabcolsep}{1pt}
For example, as previously observed in \cite{zhu2017unpaired,dhariwal2021diffusion}, the generative models do not perform geometric changes well. For the cases of pixelated pedestrians in the MOTSynth dataset as in the top row of Fig. \ref{fig:failures}, the people's shape is not reconstructed by the network. Thus, the quality of the synthetic dataset largely influences the performance of the generative model. The second example on Fig. \ref{fig:failures} is an artifact of the GAN in the sharp edges of the image objects. While this impacts the visual quality of the images, it does not necessarily worsen the performance of the detection model since GAN artifacts can act as augmentations. The bottom line of Fig. \ref{fig:failures} represents  the failure to adapt the synthetic image to the domain of MOT17: the images in MOT17 are better-lit than the night videos of MOTSynth, but here, the network failed to lighten the image, although examples from Fig. \ref{fig:inference} show that it is capable of doing so.

\section{Conclusion}
We presented a synthetic-to-real domain adaptation method for the pedestrian detection problem. While our method is based on a form of the Parametric GAN, we made important adjustments to it, specifically by changing the least squares GAN to a Wasserstein GAN and adjusting the architecture. We showcased that for the case of pedestrian detection, the reluctance of GANs towards large geometric changes is more of a blessing than a curse. This allows for the adaptation of the low-level features, while preserving geometry and thus eliminating the need of performing the adaptation of labels, which can lead to label noise and decreased performance of the downstream task. We showed that our method improves the detection accuracy when comparing to models trained exclusively on synthetic data, but also achieves performance close to the comparable models fine-tuned on the real data.  From that, we conclude that our domain adaptation method has the potential to be applicable in real setups to improve pedestrian detection if the real-world data is scarce. Moreover, since our domain adaptation method does not depend on the downstream task, it can be used as a black box for other existing models without any modifications to the loss function or architecture.
%%%%%%%%% REFERENCES
{\small
\bibliographystyle{ieee_fullname}
\bibliography{PaperForReview}
}

\end{document}